\documentclass[conference]{IEEEtran}
\IEEEoverridecommandlockouts
\usepackage{cite}
\usepackage{amsmath,amssymb,amsfonts}
\usepackage{algorithmic}
\usepackage{graphicx}
\usepackage{textcomp}
\usepackage{multirow}
\usepackage{booktabs}
\usepackage{makecell}
\usepackage{xcolor}
\usepackage{array}
\usepackage{tabularx}
\usepackage{url}
\usepackage{float}

\def\BibTeX{{\rm B\kern-.05em{\sc i\kern-.025em b}\kern-.08em
  T\kern-.1667em\lower.7ex\hbox{E}\kern-.125emX}}

\makeatletter
\def\ps@cbmsfooter{%
  \def\@oddhead{}%
  \def\@evenhead{}%
  \def\@oddfoot{%
    \hfil\footnotesize
    39th IEEE International Symposium on Computer-Based Medical Systems (CBMS 2026)%
    \hfil
  }%
  \let\@evenfoot\@oddfoot
}
\makeatother
\begin{document}

\title{Harmonized Interpretable ECG Waveform Features for Robust Cross-Dataset Clinical Prediction\\
  \thanks{\textsuperscript{*}A preliminary two-page version of this work was accepted for presentation at the 39th IEEE International Symposium on Computer-Based Medical Systems (CBMS 2026).}
}

\author{
  \IEEEauthorblockN{
    Jie Lin\IEEEauthorrefmark{1},
    Weijie Sun\IEEEauthorrefmark{2},
    Sunil V Kalmady\IEEEauthorrefmark{2},
    Anita Khalafbeigi\IEEEauthorrefmark{3}\IEEEauthorrefmark{4},
    Abram Hindle\IEEEauthorrefmark{2},
    Padma Kaul\IEEEauthorrefmark{4},
    Russell Greiner\IEEEauthorrefmark{2}\IEEEauthorrefmark{5}
  }
  \IEEEauthorblockA{
    \IEEEauthorrefmark{1}Department of Information Management,
    National Taiwan University, Taiwan
  }
  \IEEEauthorblockA{
    \IEEEauthorrefmark{2}Department of Computing Science,
    University of Alberta, Canada
  }
  \IEEEauthorblockA{
    \IEEEauthorrefmark{3}Canadian VIGOUR Center, University of Alberta, Canada}
  \IEEEauthorblockA{
    \IEEEauthorrefmark{4}Department of Medicine,
    University of Alberta, Canada}    
  \IEEEauthorblockA{    
    \IEEEauthorrefmark{5}Alberta Machine Intelligence Institute, Canada}
  \IEEEauthorblockA{
    Email: b11705048@ntu.edu.tw, weijie2@ualberta.ca, kalmady@ualberta.ca
  }
}

\maketitle
\pagestyle{cbmsfooter}

\begin{abstract}
Electrocardiograms (ECGs) are widely used for cardiovascular risk prediction, yet models often fail to transfer across hospitals because of protocol, population, and measurement differences. We benchmark cross-dataset generalization on three tasks --- heart failure classification, 30-day all-cause mortality, and 30-day mortality among sinus-rhythm ECGs --- using two large cohorts (MIMIC-IV and the Alberta Cohort). To reduce vendor-specific measurement mismatch, we build a harmonized, interpretable feature representation computed directly from raw waveforms: FeatureDB morphology/heart-rate-variability summaries plus compact time--frequency descriptors (autoregressive and wavelet features). We train XGBoost models on this unified feature space and evaluate with patient-disjoint internal and bidirectional external testing. We pre-specify two hypotheses: (H1) external AUROC retains at least 90\% of source-site internal AUROC under transfer, and (H2) internal AUROC of the harmonized feature set stays within 10\% of dataset-native machine-measurement models. 
%Across tasks, internal AUROC is 0.79--0.82 and cross-dataset AUROC is 0.74--0.78, with substantial operating-point and calibration shifts under transfer. As an exploratory benchmark, we also evaluate an end-to-end ConvNeXt model trained directly on raw ECG waveforms with age and sex; it improves absolute AUROC but shows that external validation and deployment-aware recalibration remain necessary. These findings show that a consistent waveform-derived feature interface preserves performance, supports realistic external validation, and provides a transparent alternative for cross-site clinical prediction.
Across tasks, internal AUROC is 0.79--0.82 and cross-dataset AUROC is 0.74--0.78, with larger and direction-dependent AUPRC shifts under transfer. As an exploratory benchmark, an end-to-end ConvNeXt model trained directly on raw ECG waveforms with age and sex achieves higher internal AUROC, while the harmonized representation remains competitive in relative cross-dataset transfer stability. These findings show that a consistent waveform-derived feature interface preserves performance, supports realistic external validation, and provides a transparent alternative for cross-site clinical prediction.

% \begin{itemize}
% \item We study cross-dataset generalization for clinical prediction using ECG-derived tabular features.
% \item Motivation: dataset-provided global ECG measurements are inconsistent across sources, hindering fair cross-dataset evaluation.
% \item We propose a harmonized ECG feature extraction pipeline that computes consistent handcrafted features from raw signals.
% \item Across three tasks, internal performance is competitive and cross-dataset transfer shows only modest degradation.
% \end{itemize}
\end{abstract}


\begin{figure}[t]
  \centering
  \includegraphics[width=\columnwidth]{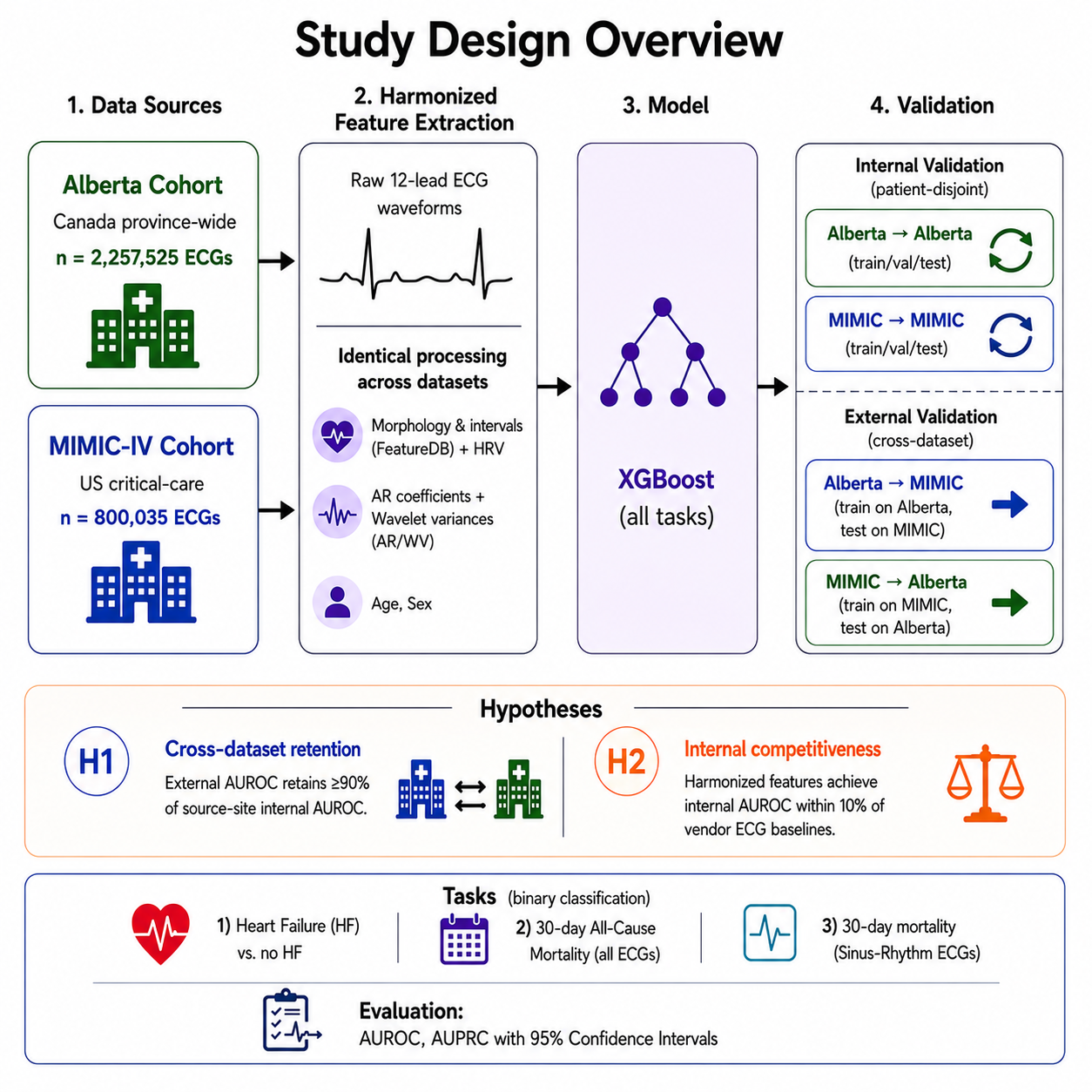}
  \caption{Overview of the study design, including the two ECG cohorts, harmonized waveform-derived feature extraction pipeline, XGBoost modeling framework, internal and bidirectional external validation settings, pre-specified hypotheses, prediction tasks, and evaluation metrics.}
  \label{fig:study_design_overview}
\end{figure}

\section{Introduction}
Machine learning electrocardiogram (ECG) models are now commonly used to predict cardiovascular risk, including heart failure \cite{Kalmady2024NPJDM} and short-term mortality~\cite{Sun2023NPJDM}. Most such models are trained and tested within a single health system, where acquisition practices, populations, and feature definitions are internally consistent. When deployed externally, shifts in these factors can degrade performance and destabilize decision thresholds. Recent evidence~\cite{leinonen2024multisource} shows that standard cross-validation and single-source evaluation can materially overestimate performance under real deployment.

Recent studies~\cite{Ribeiro2020,kalmady2024development} demonstrate strong performance from end-to-end deep learning on ECG waveforms. However, these models are hard to interpret at the feature level and remain brittle under cross-institution shift.

Many ECG datasets include high-level machine-generated measurements (such as RR interval), but these are produced by different devices, vendor-specific algorithms, and conventions. Identically named variables often encode different definitions and distributions, complicating cross-dataset evaluation and limiting transfer.

These challenges motivate two hypotheses and two research questions. H1 (cross-dataset retention): for each task and transfer direction, external AUROC will retain at least 90\% of the corresponding source-site internal AUROC (equivalently, at most 10\% relative loss). H2 (internal competitiveness): within each cohort, the harmonized FeatureDB+AR/WV model will achieve internal AUROC within 10\% of the dataset-native machine-measurement model. RQ1: what is the effect of using a harmonized representation on internal discrimination and uncertainty (95\% confidence intervals)? %RQ2: what is the effect of this harmonized representation on bidirectional external transfer performance and calibration-sensitive metrics (95\% confidence intervals)?
RQ2: what is the effect of this harmonized representation on bidirectional external transfer performance, including AUROC and AUPRC (95\% confidence intervals)?

To answer these questions, we extract tabular ECG features directly from raw waveforms and enforce identical definitions across datasets. Inputs are derived features (not full waveforms): P/QRS/T morphology and heart-rate-variability summaries, plus compact time--frequency descriptors (autoregressive (AR) and wavelet (WV) features). This design is reproducible and clinically interpretable. We evaluate on MIMIC-IV and the Alberta Cohort for three tasks: heart failure classification, 30-day all-cause mortality, and 30-day mortality among sinus-rhythm (SR) ECGs. Figure~\ref{fig:study_design_overview} summarizes the overall study design.

%To answer these questions, we extract tabular ECG features directly from raw waveforms and enforce identical definitions across datasets. Inputs are derived features (not full waveforms): P/QRS/T morphology and heart-rate-variability summaries, plus compact time--frequency descriptors (autoregressive (AR) and wavelet (WV) features). This design is reproducible and clinically interpretable. We evaluate on MIMIC-IV and the Alberta Cohort for three tasks: heart failure classification, 30-day all-cause mortality, and 30-day mortality among sinus-rhythm (SR) ECGs.

The main contributions of this study are:
  \begin{itemize}
    \item We construct a unified, waveform-derived tabular representation by adapting feature-extraction procedures from prior work (FeatureDB-style morphology/heart rate variability (HRV) features~\cite{hong2017encase,hong2019combining,fan2025detecting} plus time--frequency spectral descriptors, including AR and WV features~\cite{nahak2023evaluation}); the resulting feature set is fully reproducible from raw ECG signals and identically defined across datasets.
    \item We design a patient-disjoint evaluation protocol with internal testing and bidirectional external validation (MIMIC$\rightarrow$Alberta and Alberta$\rightarrow$MIMIC), enabling a direct assessment of cross-dataset generalization under identical feature definitions.
    \item We pre-specify hypothesis-driven evaluation criteria (AUROC retention threshold and within-cohort comparability threshold) and quantify uncertainty with 95\% bootstrap confidence intervals.
   % \item We benchmark the unified representation against each cohort's dataset-native machine measurements for internal performance, and we characterize performance and calibration shifts under cross-site transfer to inform deployment practice.
  % \item We benchmark the unified representation against each cohort's dataset-native machine measurements for internal performance, and we characterize cross-site domain shift through changes in AUROC and AUPRC to inform deployment practice.
\item We benchmark the unified representation against each cohort's dataset-native machine measurements for internal performance, and we characterize the effects of cross-site domain shift on AUROC and AUPRC to inform deployment practice.

%   \item We show that discriminative performance (AUROC/AUPRC) remains informative under transfer, while threshold-dependent metrics may drift across datasets, highlighting the need for target-site recalibration.
    \item We provide SHAP-based interpretation to connect model predictions to explicit feature groups and improve transparency for clinical review.

  \end{itemize}

As a deep learning benchmark, we train ConvNeXt models directly on raw 12-lead ECG waveforms with age and sex and evaluate them under the same internal and bidirectional external validation protocol.
% Placeholder paragraphs below: keep story clear; fill details later.
\begin{figure}[t]
  \centering
  \includegraphics[width=\columnwidth]{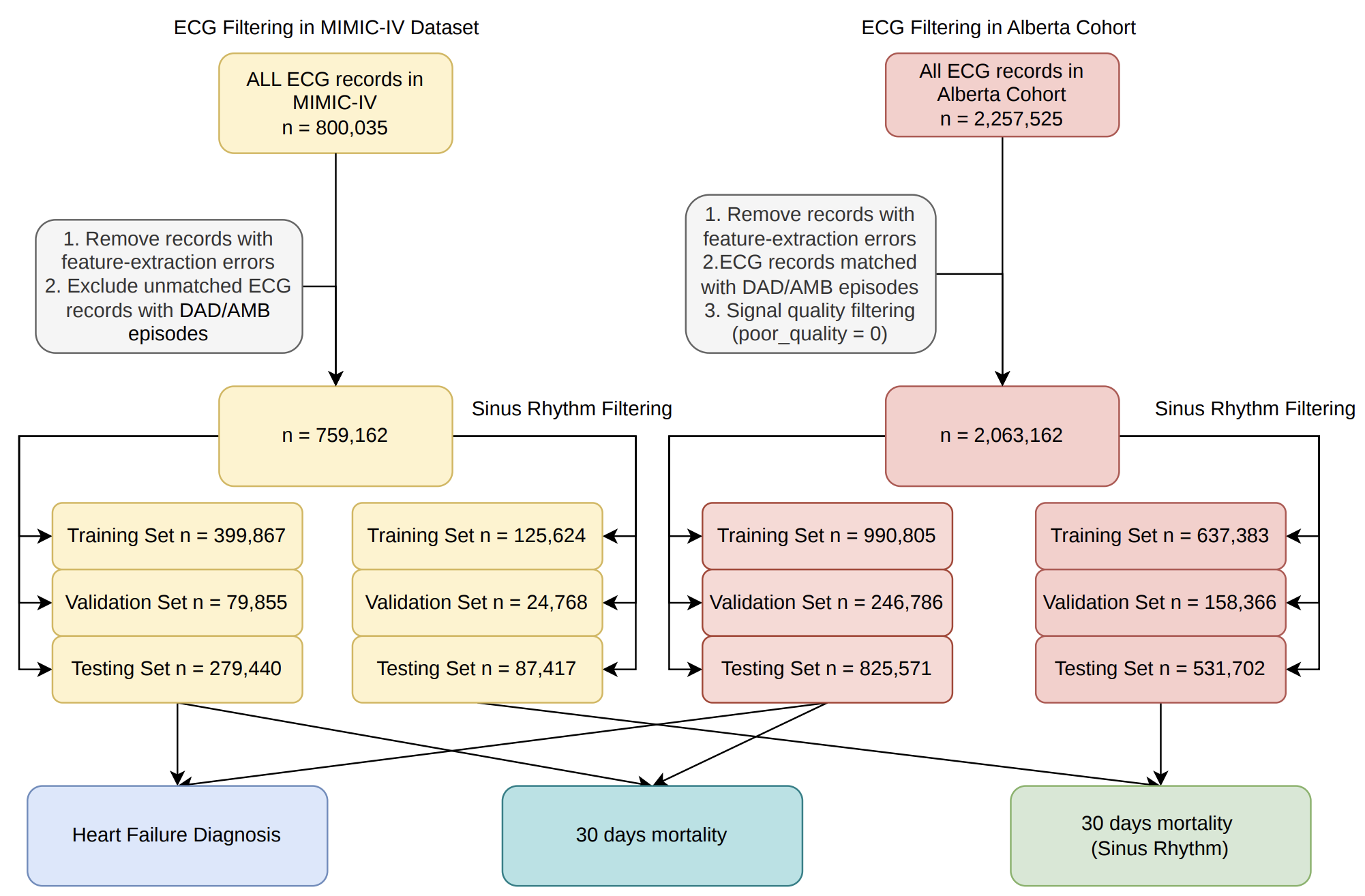}
  \caption{ECG filtering and dataset splitting (MIMIC-IV and Alberta Cohort).}
  \label{fig:flowchart}
\end{figure}

\section{Related Work}

\subsection{End-to-end deep learning on ECG}
End-to-end deep learning models on raw ECG waveforms have shown strong results in both diagnostic and prognostic tasks.
For arrhythmia detection, large-scale single-lead and 12-lead studies report clinically competitive accuracy~\cite{hannun2019cardiologist,Ribeiro2020}. Beyond diagnosis, deep ECG models have also been prospectively validated for screening low left-ventricular ejection fraction from routine ECGs, highlighting ECG's potential as a low-cost screening tool~\cite{attia2019prospective}.

Despite these successes, reviews and empirical studies~\cite{hong2020review,finlayson2021datasetshift,huang2024generalization} identify two persistent barriers to deployment: (i) cross-institution generalization under population, device, and labeling shifts, and (ii) limited feature-level interpretability, since explanations are often given at the signal-saliency level rather than via standardized ECG descriptors.

\subsection{Handcrafted ECG features and tabular models}
Handcrafted ECG features remain attractive because they align with clinical practice and provide interpretable predictors.
Classic feature families include interval and morphology measurements (e.g., PR/QRS/QT durations and wave amplitudes) and HRV indices derived from RR intervals, with measurement and interpretation guided by established standards~\cite{taskforce1996hrv,kligfield2007standardization}. In rhythm-centric benchmarks~\cite{clifford2017af,hong2019combining}, feature engineering with conventional classifiers (including gradient boosting) has shown competitive AUROC and AUPRC for atrial fibrillation detection, and hybrid approaches that integrate deep and engineered features provide further gains.

Compact time--frequency descriptors also support ECG classification.
AR modeling captures short-term dynamics for arrhythmia recognition \cite{ge2002ar}, while WV features provide multi-resolution representations that encode morphology and frequency structure \cite{qin2017wavelet}.
These lines of work motivate a unified, waveform-derived tabular interface that preserves interpretability while enabling strong tabular learners.
In this study, the unified model uses waveform-derived tabular features (interval/morphology, HRV, and AR/WV) as model inputs rather than raw waveforms or vendor-specific machine measurements in the unified feature model.

\subsection{Dataset shift and harmonized ECG measurements}
Cross-dataset deployment is challenged by dataset shift and measurement heterogeneity, where similarly named predictors arise from different acquisition, preprocessing, or measurement definitions across sources~\cite{finlayson2021datasetshift,vanCalster2019calibration,luijken2020measurement}.
Vendor-specific automated ECG algorithms can differ systematically.
Comparative studies found small but significant differences in interval measurements and recommended harmonization when comparing models or norms across systems~\cite{kligfield2018automated}.

While these issues have motivated domain adaptation and generalization methods, most operate end-to-end and do not explicitly resolve heterogeneity at the level of clinically interpretable ECG features \cite{li2025crossdomain}.

\begin{table}[t]
  \caption{Final split sizes and whole-dataset positive-class prevalence (after task-specific exclusions).}
  \label{tab:split_sizes_prevalence}
  \centering
  \footnotesize
  \setlength{\tabcolsep}{3pt}
  \renewcommand{\arraystretch}{0.95}
  \begin{tabular}{l l r r r c}
    \toprule
    \textbf{Task} & \textbf{Dataset} & \textbf{Train} & \textbf{Val} & \textbf{Test} & \textbf{Prevalence} \\
    \midrule
    HF & Alberta & 990,805 & 246,786 & 825,571 & 5.09\% \\
       & MIMIC   & \textbf{399,867}   & \textbf{79,855}  & \textbf{279,440} & 16.04\% \\
    \midrule
    30-day mortality (all) & Alberta & 990,735 & 246,758 & 825,505 & 5.71\% \\
                           & MIMIC   & \textbf{327,444}   & \textbf{65,238}  & \textbf{227,060} & 5.68\% \\
    \midrule
    30-day mortality (SR) & Alberta & 637,383 & 158,366 & 531,702 & 3.46\% \\
                          & MIMIC   & \textbf{125,624} & \textbf{24,768}  & \textbf{87,417}  & 2.68\% \\
    \bottomrule
  \end{tabular}
\end{table}

\section{Datasets and ECG Features}
\label{sec:datasets_overview}
We study cross-dataset generalization using two independent clinical ECG cohorts. MIMIC-IV~\cite{PhysioNet-mimic-iv-ecg-1.0} is a publicly available critical-care database from the United States, while the Alberta Cohort \cite{Sun2023NPJDM} is a province-wide ECG repository from a unified healthcare system in Alberta, Canada. Both contain routinely acquired 12-lead diagnostic ECGs, but differ in acquisition devices, preprocessing defaults, patient case-mix, and the definition/availability of vendor-generated measurement. To support bidirectional external validation without ad hoc feature alignment, we extract a harmonized set of interpretable tabular features directly from raw ECG waveforms and use this unified representation for all cross-dataset experiments.

% We use the MIMIC-IV-ECG matched subset (v1.0) released on PhysioNet~\cite{PhysioNet-mimic-iv-ecg-1.0}.

\subsection{Dataset-native ECG machine-derived measurements}
Each cohort provides dataset-native ECG machine measurements, which we use as within-dataset tabular baselines. These variables are generated by vendor-specific ECG software and summarize intervals, amplitudes, and other rhythm-related properties. We treat them as machine-generated outputs from each vendor pipeline and do not assume cross-vendor equivalence or near-perfect inter-system agreement (e.g., kappa $\approx 1$). Because vendors and definitions differ across cohorts, we use these measurements only for internal modeling and contextual comparison of within-dataset performance, not as a shared representation for cross-dataset evaluation.

\paragraph{MIMIC-IV machine measurements} %use machine measurements
MIMIC-IV provides a compact set of automatically computed measurements focused on cardiac timing and electrical orientation, including interval-related descriptors (e.g., RR and onset/offset markers) and frontal-plane axes for the P wave, QRS complex, and T wave. ECGs are collected on machines from multiple manufacturers (including Burdick/Spacelabs, Philips, and General Electric).  Prior to encoding, there are 9 numerical ECG measurements; we include acquisition metadata (bandwidth and filtering; categorical) and augment them with the patient’s age and sex. After one-hot encoding, there are 23 features.

\paragraph{Alberta Cohort machine measurements}
The Alberta Cohort provides a larger set of vendor-generated measurements from Philips ECG systems~\cite{philips_dxl_physicians_guide}. These measurements include heart/atrial rate, PR/QRS/QT-related intervals (and corrected QT metrics), and electrical axes in frontal and horizontal planes. Prior to encoding, there are 24 numerical measurements, augmented with age and sex (26 variables). After one-hot encoding, there are 29 features.

\subsection{Harmonized ECG feature extraction from ECG waveforms}
\label{sec:unified_features}
Because dataset-native machine measurements follow vendor-specific acquisition and processing conventions, they are not directly comparable across cohorts. Therefore, we derive a shared and reproducible representation by extracting features directly from raw ECG waveforms.

Our unified pipeline maps each ECG to a fixed-length tabular vector with identical feature definitions across datasets. FeatureDB features are extracted from Leads~II and~V5, while AR and WV are extracted from Leads~II and~V1; these lead assignments are fixed \emph{a priori}. The lead-specific statements below provide motivation for this pre-specified configuration: Leads~II and~V5 are commonly recommended for QT assessment because they provide the clearest delineation of the QT endpoint and the most stable T-wave morphology~\cite{gomez2016lqts}, while Leads~II and~V1 follow Nahak \emph{et al.} for compact time--frequency characterization~\cite{nahak2023evaluation}. Preprocessing is performed independently for each lead using the same single-lead procedure across all leads and features. All features are defined \emph{a priori}, and no data-driven feature selection is applied.

Per selected lead, FeatureDB contributes morphology/interval summaries and HRV descriptors~\cite{hong2017encase,hong2019combining,fan2025detecting}; AR/WV contributes complementary time--frequency descriptors.

\subsubsection{FeatureDB: morphology, intervals, and HRV features}
For each lead, we use a standardized single-lead pipeline to extract morphology/interval and HRV features. We (i) take a fixed-length segment from the raw waveform, (ii) apply lightweight denoising and amplitude normalization, (iii) detect R-peaks, and then derive (a) template-based morphology/interval features from an average beat aligned to the R-peak and (b) HRV features from the RR-interval series defined by the same R-peak sequence. This shared R-peak backbone keeps morphology and HRV features extraction consistent across datasets.

\paragraph{Single-lead preprocessing and R-peak detection.}
Raw waveforms are loaded and leads are identified. For each lead, we extract a 10\,s segment at a sampling rate of 500\,Hz; signals with other native rates are resampled to 500\,Hz before truncation. Preprocessing follows the FeatureDB GitHub implementation\footnote{\url{https://github.com/PKUDigitalHealth/FeatureDB}}. We then apply a short-window moving average filter (radius 3 samples; 7-point window total) and histogram-based amplitude normalization (lower saturation threshold 0.01). R-peaks are detected using a rule-based QRS detector; ECGs for which preprocessing or reliable R-peak detection fails are excluded.

\paragraph{Template-based morphology and interval features.}
After detecting R-peaks, we extract individual beats and align them to the R-peak to form an average-beat template. We delineate fiducial boundaries on the template (P onset/offset, QRS onset/offset, and T onset/offset) and derive interpretable morphology/interval descriptors such as component amplitudes and durations, RR/QT summaries, and derived QT correction measures. Beat averaging reduces sensitivity to noise and outliers while preserving physiologically meaningful waveform structure.

\paragraph{HRV features}
From the same R-peak sequence, we compute HRV descriptors summarizing rhythm dynamics across time domain, frequency domain, and nonlinear families (e.g., Poincar\'e-based measures, entropy, and fragmentation). These per-lead HRV features are concatenated with morphology/interval features to form the FeatureDB representation.

\subsubsection{AR/WV: compact time--frequency descriptors}
AR/WV features provide complementary summaries of ECG waveform structure beyond interval- and beat-level descriptors, capturing short-range temporal dependencies and frequency-localized energy that are not fully represented by time-domain morphology and HRV summaries. To complement FeatureDB, we extract compact time--frequency descriptors from Lead~II and Lead~V1 using a 10\,s segment resampled to 128\,Hz, following the work of Nahak \emph{et al.} \cite{nahak2023evaluation}. We apply a 4th-order high-pass filter with cutoff 0.5\,Hz to reduce baseline wander and a 60\,Hz notch filter ($Q=30$) to attenuate power-line interference.

\paragraph{AR features.}
We fit an AR($p$) model with $p=32$, $x_t = \sum_{k=1}^{p} a_k x_{t-k} + \varepsilon_t$, and retain the first $N_{\mathrm{AR}}=8$ coefficients $(a_1,\ldots,a_8)$. These coefficients summarize short-term temporal dependencies in the ECG waveform, and reflect local smoothness, regularity, and rhythmic structure.

\paragraph{WV features.}
We perform a stationary WV transform with Daubechies-2 at $J=4$ levels and compute wavelet-variance features as the mean squared detail coefficients at each level, $\theta_j^2 = \frac{1}{T}\sum_{t=1}^{T} d_{j,t}^2$, for $j=1,\ldots,J$.
These coefficients summarize signal energy across multiple scales.

Finally, we concatenate FeatureDB features from Leads~II and~V5 with AR/WV coefficients from Leads~II and~V1 to form the unified ECG feature vector used for downstream modeling. After one-hot encoding, the unified representation contains 373 ECG-derived features; adding age and sex yields 375 variables.%After one-hot encoding, each feature vector has a dimensionality of 375.

\section{Prediction Tasks and Evaluation Protocol}
\label{sec:tasks_eval}

\subsection{Prediction tasks}
\label{sec:tasks}
We study three binary prediction tasks in two cohorts (MIMIC-IV and Alberta Cohort) under a unified labeling and evaluation protocol. Labels are linked to the clinical episode associated with each ECG; ECGs flagged as poor quality (Alberta Cohort) and ECGs with feature-extraction failures are excluded.

Figure~\ref{fig:flowchart} summarizes the task-specific cohort filtering and patient-level dataset splits used for heart failure and mortality prediction, respectively.

% \begin{figure}[t]
%   \centering
%   \includegraphics[width=\columnwidth]{task1_flowchart.png}
%   \caption{ECG filtering and dataset splitting for the heart failure prediction task (MIMIC-IV and Alberta Cohort).}
%   \label{fig:task1_flowchart}
% \end{figure}

% \begin{figure}[t]
%   \centering
%   \includegraphics[width=\columnwidth]{task2task3_flowchart.png}
%   \caption{ECG filtering and dataset splitting for the mortality prediction tasks (MIMIC-IV and Alberta Cohort), including the sinus rhythm subset.}
%   \label{fig:task2task3_flowchart}
% \end{figure}

\paragraph{Task 1: Heart failure (HF) classification}
This task predicts \textit{HF vs.\ no HF}. The HF label is derived from the \textit{ICD-based main diagnosis code} of the clinical episode associated with each ECG record.

\paragraph{Task 2: 30-day all-cause mortality (all ECGs)}
This task predicts \textit{death within 30 days vs. survival}. In both datasets, we exclude ECGs without sufficient follow-up to determine 30-day outcome status (i.e., right-censored before 30 days without a recorded death): 33{,}563 in Alberta and 179{,}393 in MIMIC-IV.

\paragraph{Task 3: 30-day all-cause mortality (SR ECGs only)}
This task uses the same 30-day mortality label as Task~2 but restricts the cohort to sinus-rhythm ECGs to reduce rhythm-related confounding and focus on interval, morphology, and variability features. Only ECGs labeled as SR are retained in both cohorts, and right-censored ECGs without sufficient 30-day follow-up are excluded.

\subsection{Experimental design and evaluation protocol}
\label{sec:splits}
We follow a strict train/validation/test design. Models are trained on a training subset, tuned on a validation subset, then evaluated once on a disjoint holdout test split that is never used for model selection. All splits are constructed by random, patient-level sampling (with shuffling) to prevent information leakage.

We evaluate models under two settings:
\begin{itemize}
  \item \textbf{Internal validation (within-dataset):} models are trained and tuned on a dataset's training and validation sets and evaluated on its independent holdout test split.
  \item \textbf{Cross-dataset validation (bidirectional external testing):} models are developed on a source dataset (training + validation) and evaluated once on the independent holdout test split of the target dataset. We report both MIMIC$\rightarrow$Alberta and Alberta$\rightarrow$MIMIC directions.
\end{itemize}

To reduce confounding from unequal cohort sizes, we applied fair-size subsampling so that training and validation sizes match across datasets for each task. After task-specific exclusions, MIMIC was the smaller cohort for all three tasks; therefore, we downsampled the Alberta development data by random, label-stratified patient-level sampling to match the MIMIC training and validation sizes (Table~\ref{tab:split_sizes_prevalence}; \mbox{bolded entries} indicate the effective development-set sizes used).

% Table~\ref{tab:split_sizes} reports the final split sizes after task-specific exclusions.

\subsection{Class prevalence}
\label{sec:prevalence}
Table~\ref{tab:split_sizes_prevalence} reports the final train/validation/test split sizes after task-specific exclusion criteria, and the positive-class prevalence for each dataset. Prevalence is computed on all eligible ECGs after task-specific filtering, not within individual splits. All tasks are class-imbalanced, particularly SR mortality, so we report both AUROC and AUPRC.

% \begin{table*}[!t]
\begin{table*}[!htbp]
  \caption{ECG Classification Performance Across Outcomes. \\ Values are AUROC and AUPRC with 95\% confidence intervals (CIs) shown in parentheses. The final four rows report end-to-end ECG waveform ConvNeXt benchmark with age and sex.}
  \label{tab:ecg_perf_all_outcomes}
  \centering
  \footnotesize
  \setlength{\tabcolsep}{3pt}
  \renewcommand{\arraystretch}{1.25}
  \begin{tabular}{l l c c c c c c}
    \hline
    \multirow{2}{*}{\textbf{Task / Setting}}
      & \multirow{2}{*}{\textbf{Feature Set}}
      & \multicolumn{2}{c}{\textbf{Heart Failure}}
      & \multicolumn{2}{c}{\textbf{30-day Mortality (All ECGs)}}
      & \multicolumn{2}{c}{\textbf{30-day Mortality (Sinus-Rhythm)}} \\
    \cline{3-8}
      & & \textbf{AUROC} & \textbf{AUPRC}
        & \textbf{AUROC} & \textbf{AUPRC}
        & \textbf{AUROC} & \textbf{AUPRC} \\
    \hline

    \makecell[l]{Internal \\ (Alberta $\rightarrow$ Alberta)}
      & \makecell[l]{FeatureDB+ \\ AR/WV}
      & \makecell[c]{0.8166\\(0.8149--0.8184)}
      & \makecell[c]{0.1743\\(0.1716--0.1771)}
      & \makecell[c]{0.8053\\(0.8036--0.8072)}
      & \makecell[c]{0.2109\\(0.2079--0.2142)}
      & \makecell[c]{0.7933\\(0.7901--0.7960)}
      & \makecell[c]{0.1310\\(0.1273--0.1348)} \\
    \hline

    \makecell[l]{Internal \\ (MIMIC $\rightarrow$ MIMIC)}
      & \makecell[l]{FeatureDB+ \\ AR/WV}
      & \makecell[c]{0.8200\\(0.8180--0.8219)}
      & \makecell[c]{0.4564\\(0.4513--0.4612)}
      & \makecell[c]{0.8150\\(0.8116--0.8185)}
      & \makecell[c]{0.2251\\(0.2186--0.2317)}
      & \makecell[c]{0.8078\\(0.7995--0.8155)}
      & \makecell[c]{0.1119\\(0.1040--0.1221)} \\
    \hline

    \makecell[l]{Cross-dataset \\ (MIMIC $\rightarrow$ Alberta)}
      & \makecell[l]{FeatureDB+ \\ AR/WV}
      & \makecell[c]{0.7446\\(0.7427--0.7467)}
      & \makecell[c]{0.1163\\(0.1147--0.1182)}
      & \makecell[c]{0.7812\\(0.7794--0.7834)}
      & \makecell[c]{0.1732\\(0.1706--0.1760)}
      & \makecell[c]{0.7723\\(0.7690--0.7753)}
      & \makecell[c]{0.1081\\(0.1053--0.1111)} \\
    \hline

    \makecell[l]{Cross-dataset \\ (Alberta $\rightarrow$ MIMIC)}
      & \makecell[l]{FeatureDB+ \\ AR/WV}
      & \makecell[c]{0.7714\\(0.7691--0.7736)}
      & \makecell[c]{0.3559\\(0.3515--0.3601)}
      & \makecell[c]{0.7736\\(0.7699--0.7775)}
      & \makecell[c]{0.1692\\(0.1648--0.1742)}
      & \makecell[c]{0.7810\\(0.7724--0.7895)}
      & \makecell[c]{0.0898\\(0.0839--0.0979)} \\
    \hline

    \makecell[l]{Internal \\ (Alberta $\rightarrow$ Alberta)}
      & \makecell[l]{Machine-Derived \\ ECG Measurements}
      & \makecell[c]{0.8318\\(0.8301--0.8335)}
      & \makecell[c]{0.1932\\(0.1906--0.1963)}
      & \makecell[c]{0.8060\\(0.8043--0.8079)}
      & \makecell[c]{0.2022\\(0.1992--0.2052)}
      & \makecell[c]{0.7963\\(0.7938--0.7994)}
      & \makecell[c]{0.1326\\(0.1295--0.1364)} \\
    \hline

    \makecell[l]{Internal \\ (MIMIC $\rightarrow$ MIMIC)}
      & \makecell[l]{Machine-Derived \\ ECG Measurements}
      & \makecell[c]{0.8215\\(0.8194--0.8235)}
      & \makecell[c]{0.4598\\(0.4549--0.4648)}
      & \makecell[c]{0.7951\\(0.7914--0.7987)}
      & \makecell[c]{0.1940\\(0.1887--0.1998)}
      & \makecell[c]{0.7879\\(0.7794--0.7962)}
      & \makecell[c]{0.1042\\(0.0960--0.1144)} \\

        \hline
    \makecell[l]{Internal \\ (Alberta $\rightarrow$ Alberta)}
      & \makecell[l]{ECG waveform \\ ConvNeXt}
      & \makecell[c]{0.8779\\(0.8765--0.8792)}
      & \makecell[c]{0.2859\\(0.2817--0.2903)}
      & \makecell[c]{0.8471\\(0.8454--0.8488)}
      & \makecell[c]{0.2648\\(0.2612--0.2688)}
      & \makecell[c]{0.8353\\(0.8326--0.8380)}
      & \makecell[c]{0.1720\\(0.1674--0.1768)} \\
    \hline

    \makecell[l]{Internal \\ (MIMIC $\rightarrow$ MIMIC)}
      & \makecell[l]{ECG waveform \\ ConvNeXt}
      & \makecell[c]{0.8657\\(0.8641--0.8673)}
      & \makecell[c]{0.5477\\(0.5427--0.5522)}
      & \makecell[c]{0.8453\\(0.8422--0.8485)}
      & \makecell[c]{0.2541\\(0.2476--0.2616)}
      & \makecell[c]{0.8309\\(0.8239--0.8379)}
      & \makecell[c]{0.1256\\(0.1169--0.1347)} \\
    \hline

    \makecell[l]{Cross-dataset \\ (MIMIC $\rightarrow$ Alberta)}
      & \makecell[l]{ECG waveform \\ ConvNeXt}
      & \makecell[c]{0.8153\\(0.8136--0.8170)}
      & \makecell[c]{0.1700\\(0.1675--0.1726)}
      & \makecell[c]{0.7939\\(0.7920--0.7958)}
      & \makecell[c]{0.1920\\(0.1891--0.1949)}
      & \makecell[c]{0.7783\\(0.7750--0.7817)}
      & \makecell[c]{0.1122\\(0.1093--0.1156)} \\
    \hline

    \makecell[l]{Cross-dataset \\ (Alberta $\rightarrow$ MIMIC)}
      & \makecell[l]{ECG waveform \\ ConvNeXt}
      & \makecell[c]{0.8252\\(0.8233--0.8271)}
      & \makecell[c]{0.4655\\(0.4602--0.4704)}
      & \makecell[c]{0.8296\\(0.8264--0.8330)}
      & \makecell[c]{0.2387\\(0.2316--0.2453)}
      & \makecell[c]{0.8065\\(0.7983--0.8147)}
      & \makecell[c]{0.1185\\(0.1095--0.1285)} \\
    \hline
  \end{tabular}
\end{table*}

\subsection{Evaluation metrics}
\label{sec:metrics}
Performance is evaluated only on the independent holdout test sets. Primary metrics are AUROC and AUPRC, which summarize threshold-agnostic discrimination across decision thresholds. To test H1, we compute AUROC retention in each transfer direction as $R_{\mathrm{AUROC}}=\mathrm{AUROC}_{\mathrm{external}}/\mathrm{AUROC}_{\mathrm{internal,source}}$ and consider H1 supported when $R_{\mathrm{AUROC}} \ge 0.90$ (equivalently, relative AUROC loss $\le$10\%). Because this study does not define task-specific clinical utility weights for false positives and false negatives, we focus on discrimination and report uncertainty via 95\% confidence intervals computed from 1{,}000 non-parametric bootstrap resamples of each test set (sampled with replacement). To answer RQ1 and RQ2, we also report internal-vs-external shifts in point estimates using $\Delta$AUROC and $\Delta$AUPRC, interpreted alongside 95\% CIs for AUROC and AUPRC.

% \FloatBarrier

\section{Methods}
\subsection{Model}
We use Extreme Gradient Boosting (XGBoost)~\cite{chen2016xgboost} as the core prediction model for all tasks and experimental settings. We train separate binary classifiers using either (i) the harmonized feature set extracted from raw ECG waveforms (FeatureDB from Leads~II and~V5; AR/WV from Leads~II and~V1) or (ii) the dataset-native machine ECG measurements (within-dataset baselines). We treat each ECG as an independent sample and use patient-level train/validation/test splits as described in Section~\ref{sec:splits}. Hyperparameters are selected on the validation split, and final performance is reported on the held-out set.

Before training, we apply 99\% winsorization to numerical features to reduce the influence of extreme outliers~\cite{wilcox2017modern}. Missing numerical values, mainly from unavailable or low-quality waveform segments, are imputed with training-set means, and missing categorical values are assigned to a dedicated \texttt{nan} category. We use this simple imputation strategy for consistency across cohorts, and note that it does not model the missingness mechanism explicitly~\cite{little2019statistical}.

We chose XGBoost because it performs well on heterogeneous tabular clinical features, captures nonlinear interactions, and supports post-hoc interpretation through standard feature-attribution methods such as SHAP~\cite{NIPS2017_7062}.

\subsection{Hyperparameter optimization}
\label{sec:hparam_ranges}
Hyperparameter optimization is performed with Optuna~\cite{akiba2019optuna} (sequential model-based optimization). For each task and dataset, we run an independent Optuna study with 30 trials and select the configuration that maximizes validation AUROC.

We tune XGBoost hyperparameters separately for Alberta Cohort and MIMIC using each dataset's training/validation split. The search space includes learning rate (0.001--0.2), maximum tree depth (2--10), minimum child weight (0.01--50), instance subsampling (0.5--1.0), feature subsampling per tree (0.5--1.0), minimum loss reduction for a split (0--10), and L1/L2 regularization ($10^{-9}$--10; $10^{-9}$--50). Log-uniform sampling is used where appropriate.

For each trial, we train the models up to 5{,}000 boosting iterations with early stopping if validation AUROC does not improve for 200 consecutive rounds. After optimization, we refit the model using the selected configuration and evaluate once on the held-out set.

\begin{figure*}[!htbp]
\centering
\includegraphics[width=\textwidth,height=0.55\textheight,keepaspectratio]{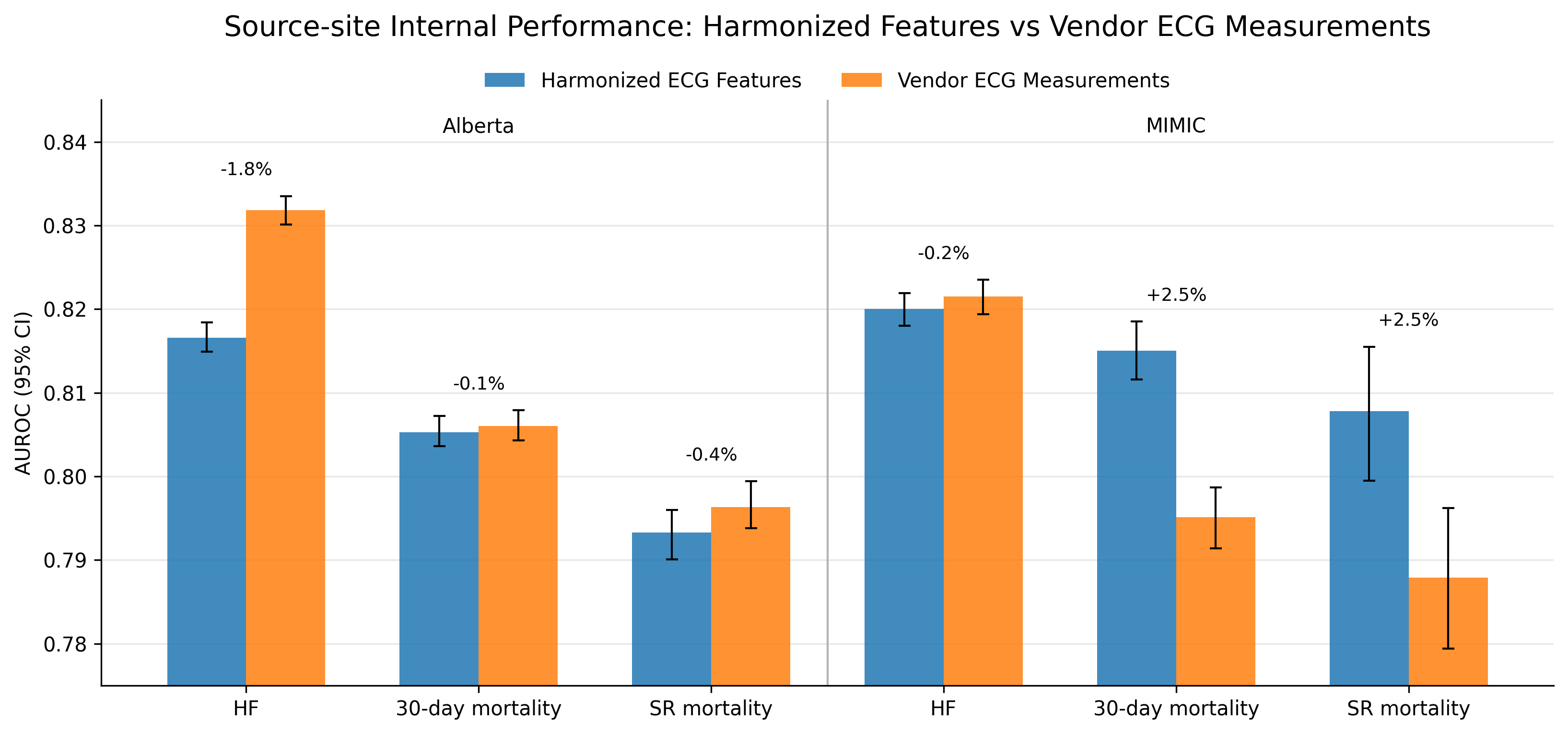}
\caption{Source-site internal AUROC comparison between the harmonized FeatureDB+AR/WV representation and dataset-native machine-derived ECG measurements. Error bars show 95\% confidence intervals. Percent labels indicate the relative AUROC difference of FeatureDB+AR/WV compared with the corresponding machine-measurement baseline.}
\label{fig:h2_internal_harmonized_vs_vendor}
\end{figure*}

\subsection{Deep learning waveform benchmark}
%As a deep learning waveform benchmark, we trained a ConvNeXt-based end-to-end model using raw 12-lead ECG waveforms with a sequence length of 4096 samples, together with age and sex. Models were trained separately for each task and source dataset using the same patient-level splits and internal/cross-dataset evaluation protocol as the FeatureDB+AR/WV models. Performance was summarized using AUROC and AUPRC with 95\% bootstrap confidence intervals.

As a deep learning waveform benchmark, we trained a one-dimensional ConvNeXt V2 model~\cite{Woo2023ConvNeXtV2} on raw 12-lead ECG waveforms with 4096 samples per lead, together with age and sex. Models were trained using binary cross-entropy loss and the AdamW optimizer, with early stopping based on validation loss. Separate models were trained for each task and source dataset using the same patient-level splits and internal/cross-dataset evaluation protocol as the FeatureDB+AR/WV models.

\section{Experimental Results}

Table~\ref{tab:ecg_perf_all_outcomes} summarizes AUROC/AUPRC (95\% CIs) for three tasks under (i) internal testing within each cohort and (ii) bidirectional external validation (MIMIC$\rightarrow$Alberta, Alberta$\rightarrow$MIMIC). Results are reported primarily for the unified FeatureDB+AR/WV representation; dataset-native machine-derived ECG measurements are included as internal tabular baselines, and the ECG waveform ConvNeXt model is included as an exploratory end-to-end benchmark. Overall, FeatureDB+AR/WV tracks machine-derived baselines internally, while external transfer shows modest AUROC declines but larger AUPRC shifts, consistent with prevalence differences and broader cross-site domain shift.

%Hypothesis-oriented summary: H1 is supported in all six FeatureDB+AR/WV transfer evaluations. AUROC retention ranges from 0.908 to 0.985 (HF: 0.908 and 0.945; 30-day mortality all ECGs: 0.959 and 0.961; 30-day mortality SR: 0.956 and 0.985), meeting the pre-specified $\ge$0.90 criterion in every direction. H2 is also supported: internal AUROC of the harmonized representation remains within 10\% of machine-measurement baselines across all dataset-task pairs, with relative differences within approximately $\pm$2.5\%.
%These results indicate that harmonization preserves within-cohort discrimination while maintaining strong transfer discrimination across sites.

Hypothesis-oriented summary: H1 is supported in all six FeatureDB+AR/WV transfer evaluations. AUROC retention ranges from 0.908 to 0.985 (HF: 0.908 and 0.945; 30-day mortality all ECGs: 0.959 and 0.961; 30-day mortality SR: 0.956 and 0.985), meeting the pre-specified $\ge$0.90 criterion in every direction. H2 is also supported: internal AUROC of the harmonized representation remains within 10\% of machine-measurement baselines across all dataset-task pairs, with relative differences within approximately $\pm$2.5\% (Figure~\ref{fig:h2_internal_harmonized_vs_vendor}).
These results indicate that harmonization preserves within-cohort discrimination while maintaining strong transfer discrimination across sites.

\begin{figure*}[t]
\centering
\includegraphics[width=\textwidth,height=0.55\textheight,keepaspectratio]{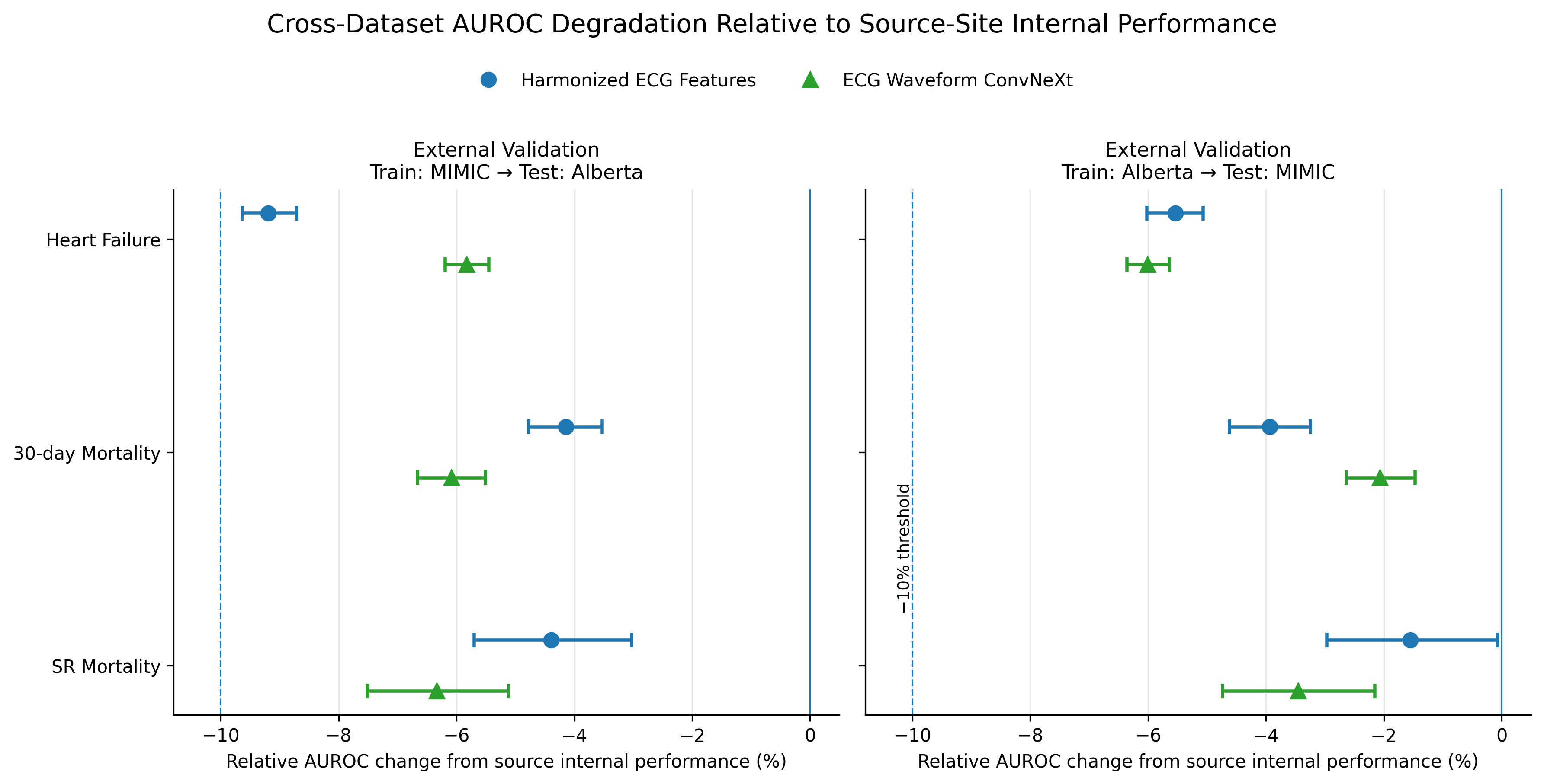}
\caption{Relative external AUROC change from source-site internal performance for the main harmonized FeatureDB+AR/WV model and the end-to-end ECG waveform ConvNeXt benchmark. Negative values indicate lower AUROC under external validation. The dashed vertical line marks the pre-specified 10\% relative AUROC loss threshold used for H1.}
\label{fig:auroc_degradation}
\end{figure*}

\textbf{HF classification:}
Internal FeatureDB+AR/WV performance is competitive with machine-derived measurements (Alberta AUROC 0.8166 vs 0.8318; MIMIC AUROC 0.8200 vs 0.8215), and AUPRCs are in the same range (Table~\ref{tab:ecg_perf_all_outcomes}). External validation shows the largest prevalence-sensitive shifts: MIMIC$\rightarrow$Alberta reduces AUROC by 0.075 and AUPRC by 0.340, while Alberta$\rightarrow$MIMIC reduces AUROC by 0.045 but increases AUPRC by 0.182 relative to internal FeatureDB+AR/WV, indicating domain shift under transfer.

\textbf{30-day mortality (all ECGs):} 
Internal FeatureDB+AR/WV performance is on par with or slightly above machine-derived measurements (MIMIC AUROC 0.8150 vs 0.7951; Alberta AUROC 0.8053 vs 0.8060). Under transfer, AUROC reductions are modest (0.032--0.034) and AUPRC drops are moderate (0.042--0.052), consistent with relatively stable discrimination despite domain shift under transfer.

\textbf{30-day mortality (SR):}
Within SR ECGs, internal FeatureDB+AR/WV remains close to machine-derived baselines (Alberta AUROC 0.7933 vs 0.7963; MIMIC AUROC 0.8078 vs 0.7879). Cross-dataset AUROC drops are small (0.012--0.036), while AUPRC shifts are asymmetric (0.004 to 0.041), again consistent with prevalence differences and domain shift between cohorts.

The ConvNeXt waveform model achieved higher absolute AUROC than FeatureDB+AR/WV across internal and external settings (Table~\ref{tab:ecg_perf_all_outcomes}). This gain should be interpreted in light of the larger input dimensionality: with a sequence length of 4096 samples, ConvNeXt used 49{,}152 raw waveform input values per ECG plus age and sex, whereas FeatureDB+AR/WV used 375 interpretable variables and the machine-measurement baselines used 23--29 variables. However, ConvNeXt showed the same deployment pattern as the harmonized-feature model: external AUROC was lower than source-site internal AUROC, and AUPRC shifted across datasets. Figure~\ref{fig:auroc_degradation} summarizes the relative AUROC change under transfer for both models, showing that both remained within the 10\% relative AUROC loss threshold while still highlighting the need for external validation and, where appropriate, deployment-aware domain adaptation. Taken together, the ConvNeXt experiments demonstrate stronger absolute discrimination, whereas the harmonized representation retained competetive relative robustness under cross-dataset transfer.

\begin{figure*}[t]
  \centering
  \includegraphics[width=\textwidth,height=0.75\textheight,keepaspectratio]{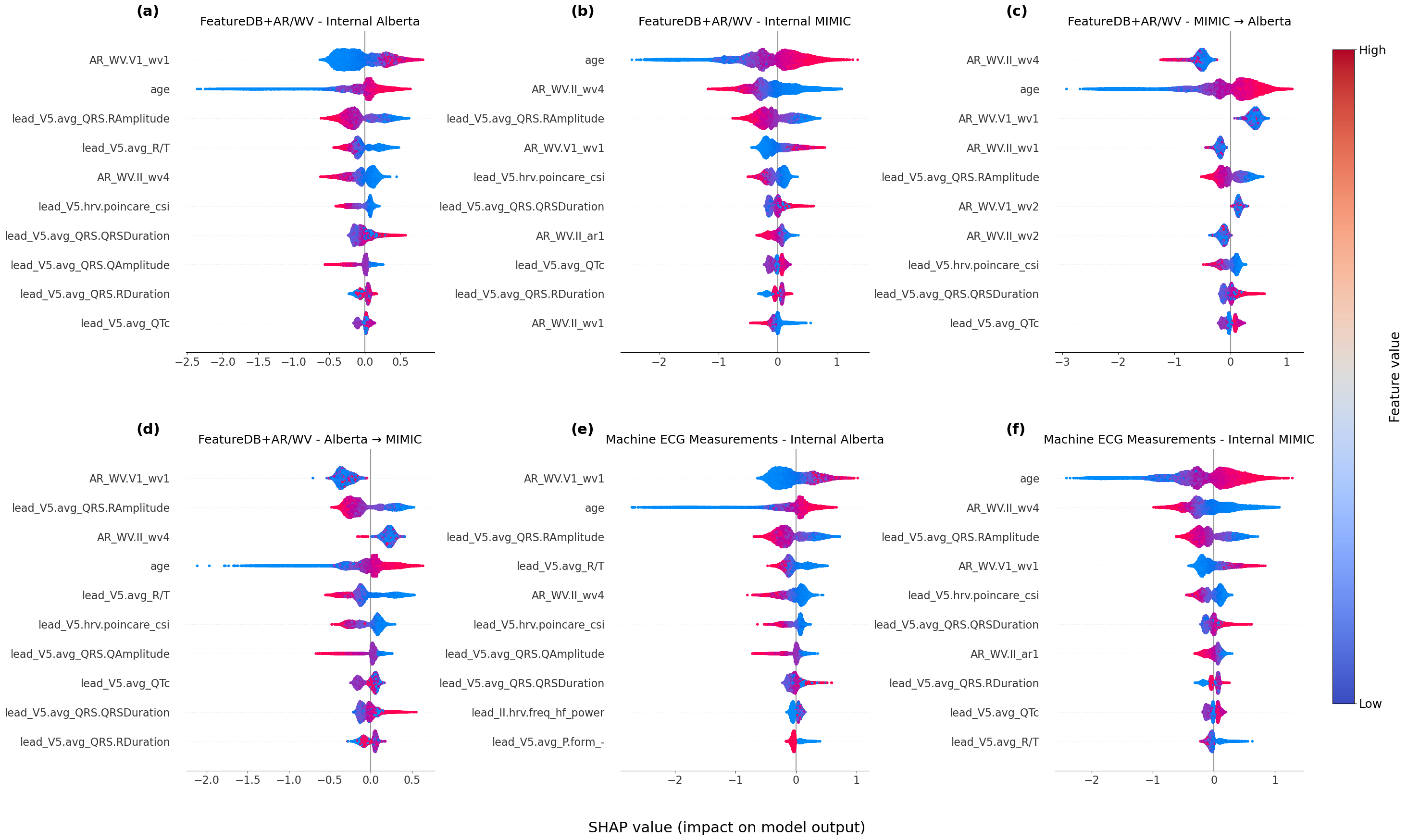}
  \caption{SHAP summary plots for Task~1 (heart failure classification) across internal evaluation and cross-dataset transfer settings. Only the top 10 features ranked by mean absolute SHAP value are shown for clarity.}
  \label{fig:shap_summary}
\end{figure*}

\section{Model Interpretability (SHAP)}

\subsection{Setup}
We use Shapley Additive Explanations (SHAP) \cite{NIPS2017_7062} to attribute model predictions to individual input features and to verify that the learned decision rules align with physiologically plausible ECG signatures. To examine the robustness and interpretability of the unified feature space, we compute SHAP for XGBoost models trained on FeatureDB+AR/WV and on dataset-native machine ECG measurements, and summarize feature importance by mean absolute SHAP value.

Due to space constraints, we report SHAP analyses for Task~1 (HF classification) only. SHAP is computed on the held-out set of each corresponding evaluation setting; when computational constraints apply, we use a stratified subsample with 25\% positives and 75\% negatives to reduce computation while preserving a consistent class balance.

\subsection{Key observations}
Figure~\ref{fig:shap_summary} highlights stable drivers of Task~1 across internal and transfer settings. In the FeatureDB+AR/WV models, the dominant contributors consistently include age, AR/WV time--frequency terms (e.g., AR\_WV.V1\_wv1 and AR\_WV.II\_wv4), and Lead~V5 morphology (e.g., QRS R-amplitude, QRS duration, and R/T ratio). The color gradients are largely monotonic, with higher age and larger AR/WV values typically pushing predictions toward higher HF risk, while morphology/interval terms provide complementary physiological structure.

Under transfer, importance shifts toward AR/WV terms (panels c--d), while several morphology and HRV features show reduced spread, suggesting that waveform-derived time--frequency descriptors are more stable across cohort and device shifts. This pattern is consistent with the idea that AR/WV terms capture shape-based signatures that are less sensitive to vendor-specific measurement conventions.

Compared with models trained on vendor machine measurements (panels e--f), the unified feature space yields a more balanced attribution profile that mixes AR/WV, morphology, and HRV signals rather than concentrating on a small set of vendor-defined global measurements. This distribution supports more comparable interpretation and auditing across sites.

\section{Discussion and Conclusion}
This study targets a practical deployment gap in ECG-based ML: models trained on a single health system often fail to transfer because measurement definitions, devices, and populations differ. End-to-end waveform models can be accurate but are harder to audit and are sensitive to acquisition and preprocessing changes. At the same time, vendor-provided machine measurements are often inconsistent across datasets, limiting fair external validation. Our core contribution is a harmonized, waveform-derived feature interface (FeatureDB morphology/HRV plus AR and WV) that is identically defined across cohorts and preserves clinical interpretability.

Across three tasks and two datasets, the unified features achieve internal performance comparable to dataset-native machine measurements (Table~\ref{tab:ecg_perf_all_outcomes}), suggesting that most predictive signal in vendor measurements is recoverable from transparent, reproducible extraction. In response to our hypotheses, H1 is supported because external AUROC remains within 10\% of source internal AUROC in all transfer directions, and H2 is supported because internal AUROC with harmonized features remains within 10\% of machine-measurement models across all tasks. With identical feature definitions, bidirectional external validation showed modest AUROC declines of approximately 0.02--0.08, but larger AUPRC changes--up to 0.34 for heart failure, 0.05 for all-ECG mortality, and 0.04 for sinus-rhythm mortality. These shifts likely reflect prevalence differences and broader domain shift. Thus, AUROC was relatively preserved, whereas precision--recall performance was more sensitive to the target population, highlighting the need for target-site evaluation, recalibration, threshold adjustment, and, where needed, domain adaptation before deployment.

These findings support harmonized handcrafted ECG features as a dataset-independent interface for external validation, auditing, and consistent interpretation across institutions. This approach may reduce feature re-engineering during transfer and make site-specific adaptation needs clearer. Limitations include evaluation on only two cohorts, a limited set of leads and feature families, and no assessment of recalibration, domain adaptation, subgroup transportability, or clinically defined decision thresholds. Future work should expand validation across vendors and health systems and assess broader feature coverage, subgroup performance, adaptation methods, cost-sensitive thresholds, and post-deployment monitoring.

\bibliography{references_2}

\end{document}